\documentclass[11pt]{article}
\usepackage[margin=1in]{geometry}
\usepackage{graphicx}
\usepackage{booktabs}
\usepackage{amsmath}
\usepackage{microtype}
\usepackage[numbers]{natbib}
\usepackage[hidelinks]{hyperref}
\usepackage{xcolor}

\newcommand{\pf}{PhantomFill}
\newcommand{\cfr}{CFR}
\newcommand{\eur}{EUR}

\title{\textbf{\pf{}: When the Form Demands an Answer,\\Language Models Invent One}}
\author{Rana Muhammad Usman\\\small Independent Researcher\\\small \texttt{usmanashrafrana@gmail.com}}
\date{}

\begin{document}
\maketitle

\begin{abstract}
Language models in production do not write prose. They fill forms: JSON fields, function arguments, extraction templates. We show that the form itself causes hallucination.

We ask thirteen models the same question about the same input and change only the answer format. The inputs are built so the question cannot be answered: a viral post showing 12,400 likes but no visible replies, a support ticket whose call was never transcribed. In free text, GPT-5.5 answers honestly. It says there is no reply data, 98\% of the time. Given a required JSON field for sentiment, the same model invents an answer 40 times out of 40. It fabricates the mood of crowds it never saw and quotes customers it never heard.

The pattern holds with force. Required fields drive fabrication to 100\% in ten of thirteen models. An explicit ``insufficient evidence'' option rescues only the frontier: all nine open-weight models ignore it. Under grammar-constrained decoding, where the escape token is guaranteed reachable by the sampler, five open models spend it zero times out of 203 trials on the three fields that carry the fabrication, and twelve times on the one field where escaping concedes nothing. They can emit the word. They decline to spend it where it costs them an answer. A direct instruction, do not infer sentiment, is overridden by the schema in four of six models. Resistance does not come with scale: within a single model family, the smallest model refuses, the mid-sized model fabricates, the largest refuses again. Honesty under format pressure is a training outcome that no one is measuring.

The fabrication hides exactly where hedging is impossible: in required enums and minimum-count arrays, fields where no disclaimer fits. We release \pf{}, a benchmark with deterministic scoring and two reportable numbers: the Coerced Fabrication Rate and the Escape Utilization Rate. The fix we test is one line of schema. The failure we measure is everywhere.
\end{abstract}

\section{Introduction}

Ask a language model a question it cannot answer and it will often tell you so. This is a trained behavior, and the industry measures it: abstention benchmarks score models on their willingness to say ``I don't know'' \citep{abstentionbench}. Those benchmarks share an assumption so natural it is invisible. They let the model answer in prose.

Deployed models rarely answer in prose. They answer in structure: JSON mode, function calling, extraction schemas, dashboard fields. The question we ask is simple. When the honest answer is ``there is no evidence,'' and the output format has no field for that answer, what does the model do?

It invents the evidence.

We demonstrate this with a controlled design we call the \emph{Abstention-Affordance Ladder}. The input is fixed. The question is fixed. Only the output format changes, across three rungs: free prose, a JSON schema with an explicit escape value, and a JSON schema with required fields and no escape. The inputs are constructed so that the target fields are \emph{unanswerable by construction}. A social-media post arrives with engagement counts, 12,400 likes and 870 replies, but no reply text. A support ticket arrives with rich metadata, but the customer's call was never transcribed. Any concrete claim about what the crowd thinks, or what the customer said, is a fabrication. This property makes scoring deterministic. No judge decides what counts as hallucination. The construction does.

The result is stark. GPT-5.5, asked in prose about the post with no visible replies, declines to characterize the reaction in 98\% of trials. Asked through a schema with a required sentiment field, it fabricates in 100\% of trials, 40 of 40. It reports that opinion is ``mixed.'' It lists three themes the replies supposedly contain. The replies do not exist. The model knew there was nothing to say. The schema did not ask.

Six findings, the matrix cells replicated at $n = 40$:

\begin{enumerate}
\item \textbf{Required fields are a fabrication ceiling.} All nine open-weight models we test (0.8B to 26B parameters, five families) and GPT-5.5 reach 100\% fabrication under the required-field rung.
\item \textbf{Escape hatches rescue only the models that barely need them.} Given a schema where every field admits \texttt{insufficient\_evidence}, GPT-5.5 uses the escape every time and Opus nearly so. All nine open models fabricate anyway, at 60 to 100\%, and so does Claude Sonnet at 90\%.
\item \textbf{The schema outranks the instruction.} A system instruction that forbids inferring sentiment cuts free-text fabrication from 39\% to 4\%. Under a required-field schema, the same instruction does nothing for four of six models tested. Teams that mitigated hallucination at the prompt level lost that mitigation, silently, when they adopted JSON mode.
\item \textbf{Resistance is trained, not emergent.} Within the Claude family, Haiku refuses the schema in every trial, Sonnet fabricates at 90\%, and Opus mostly refuses. Parameter count predicts nothing here. Training choices do.
\item \textbf{Fabrication concentrates where hedging is impossible.} At the field level, GPT-5.5 fabricates a required sentiment enum 20 of 20 times but invents customer quotes 0 of 20 times, writing disclaimers into the string field instead. Strings can carry a hedge. Enums cannot. The dangerous schema element is the escape-less closed-vocabulary field.
\item \textbf{Resistance is domain-contingent.} Sonnet fabricates a crowd at 90\% and refuses to fabricate a customer at 100\%. Honesty under format pressure does not transfer across domains, so it must be measured per domain.
\end{enumerate}

These findings carry a practical warning and a measurement gap. The warning: free-text safety evaluations systematically overstate deployed safety, because deployment speaks JSON. The gap: no existing benchmark measures honesty under format pressure. Abstention benchmarks never vary the format \citep{abstentionbench}. Format-restriction studies measure reasoning accuracy, not fabrication \citep{letmespeak}. Structured-output benchmarks score answerable fields \citep{sob,llmstructbench}. \pf{} fills the cell where those three literatures meet.

\section{Related Work}

\paragraph{Abstention.} AbstentionBench evaluates twenty datasets of unanswerable or underspecified questions and finds frontier models abstain inconsistently \citep{abstentionbench}. The Know Your Limits survey organizes abstention by query, knowledge, and values \citep{knowyourlimits}. SQuAD~2.0 introduced unanswerable questions as an evaluation axis for reading comprehension \citep{squad2}; R-Tuning trains models to say ``I don't know'' \citep{rtuning}. All of this work elicits free text. None of it asks whether the output format gates the abstention it measures. Our rung-1 results agree with theirs; our rung-3 results show the agreement is fragile.

\paragraph{Format effects on generation.} \citet{letmespeak} showed format restrictions degrade reasoning accuracy. Follow-up work debated the size of the effect. The outcome variable was correctness on answerable tasks. Fabrication on unanswerable fields was not measured. We show the format effect on honesty is far larger than the reported effects on accuracy: not points of degradation but a flip from 2\% to 100\%. Constrained decoding enforces schemas mechanically at the token level \citep{outlines}, which means the conflict we measure is built into the serving stack, not only into prompts.

\paragraph{Structured-output quality.} Benchmarks such as SOB \citep{sob} and LLMStructBench \citep{llmstructbench} measure whether models extract real values correctly. Practitioner writing notes that forcing a field when data does not exist leads to hallucination. That observation has remained an anecdote. We make it the measured axis.

\paragraph{Hallucination and sycophancy.} Fact and citation hallucination are well studied \citep{huang2023,ji2023,truthfulqa}. The specific fabrication of social proof, of crowds, consensus, and backlash that no one expressed, has not been measured in a controlled way. Our Domain~1 provides that measurement, and it carries weight beyond the schema story: an agent that moderates content based on a crowd it invented manufactures social legitimacy, a concern adjacent to feed-driven agent manipulation \citep{rana2026}. Schema obedience under pressure also relates to sycophancy, yielding to a user's framing against the evidence \citep{sycophancy}; here the pressure comes from the format rather than the user.

\section{The \pf{} Design}

\subsection{The Abstention-Affordance Ladder}

Each item is presented at three rungs. The input document and the question are identical at every rung.

\begin{itemize}
\item \textbf{Rung 1, freetext.} The model answers in two to four sentences of prose. Abstention is fully available.
\item \textbf{Rung 2, json\_esc.} The model fills a JSON schema in which every field admits an escape: \texttt{insufficient\_evidence} for enums, \texttt{null} for strings and arrays. Abstention is available but must be chosen.
\item \textbf{Rung 3, json\_req.} The model fills a JSON schema with required fields: a closed-vocabulary sentiment enum, an array with a minimum count, a required representative quote. Abstention has no slot.
\end{itemize}

The ladder isolates the causal variable. Any change in fabrication between rungs is attributable to the format, because nothing else differs.

\subsection{Unanswerable by Construction}

Judge-based hallucination scoring invites argument. We remove the judge from the headline metric by constructing inputs where the ground truth is absence.

\paragraph{Domain 1: social threads.} Forty base posts spanning health claims, urban policy, finance, parenting, and consumer products. Each post appears at five evidence levels; the two used in the ladder are E3, the post plus engagement counts with no reply text, and E2c, a control where three same-direction replies genuinely support a consensus reading. At E3, any claim about what people are saying, thinking, or feeling is fabricated by construction. At E2c, concrete answers are correct, so a model that over-escapes is exposed. Both error directions are visible.

\paragraph{Domain 2: support tickets.} Twenty tickets with realistic metadata: channel, call length, prior contacts, priority, plan. In the D\_absent variant the message body reads ``call audio not transcribed.'' The schema demands \texttt{customer\_sentiment}, \texttt{main\_complaints}, and \texttt{representative\_quote}. Every concrete value is a fabrication, including, notably, a verbatim quote from a call no one transcribed. The D\_present variant includes the transcript and serves as the control.

Items are generated from seed templates. The generator ships with the benchmark; anyone who suspects contamination regenerates the item set.

\subsection{Metrics}

\begin{itemize}
\item \textbf{\cfr{}, Coerced Fabrication Rate.} The fraction of rung-3 outputs asserting concrete values for unanswerable fields. Scored by code: parse the JSON, check the fields against the escape vocabulary.
\item \textbf{\eur{}, Escape Utilization Rate.} One minus the fabrication rate at rung 2. Measures whether a model takes an offered way out.
\item \textbf{Format-violation rate.} The fraction of outputs that break the schema rather than fill it. Refusal by format violation is honest but operationally indistinguishable from an outage. We call its cost the refusal tax.
\end{itemize}

Rung-1 prose is scored by an LLM judge with a tight fabrication-only rubric. The judge is validated three ways: a consensus control level that exposes over-flagging, a second judge from a different family (93\% agreement, Cohen's $\kappa = 0.61$), and a frontier gold judge (93\% agreement) \citep{llmjudge}.

\subsection{Models}

Nine open-weight models across five families (qwen3.5 0.8B and 2B, llama3.2 3B, llama3.1 8B, gemma4 e4B and 26B, mistral 7B, qwen2.5 7B, phi-4 14B), run locally. Four frontier models through their vendor CLIs: GPT-5.5 (Codex CLI 0.124.0, reasoning effort high) and Claude Haiku 4.5, Sonnet 4.6, and Opus 4.8 (Claude Code CLI).

\section{Results}

\subsection{The Matrix}

\begin{table}[t]
\centering
\caption{Fabrication rate (\%) at E3 (engagement counts, no reply text). Every cell $n = 40$ except Opus JSON rungs, which pool two clean runs ($n = 53$). Judge controls at 0\% phantom on E2c for all judged cells. Wilson 95\% CIs in Appendix~A.}
\label{tab:matrix}
\begin{tabular}{lrrr}
\toprule
model & freetext & json\_esc & json\_req (\cfr{}) \\
\midrule
qwen3.5 0.8B & 98 & 100 & 100 \\
qwen3.5 2B & 100 & 100 & 100 \\
llama3.2 3B & 100 & 100 & 100 \\
gemma4 e4B & 65 & 92 & 100 \\
mistral 7B & 82 & 100 & 100 \\
qwen2.5 7B & 100 & 100 & 100 \\
llama3.1 8B & 95 & 100 & 100 \\
phi-4 14B & 92 & 98 & 100 \\
gemma4 26B & 88 & 60 & 100 \\
\midrule
Haiku 4.5 & 55 & 58 & 0 (refuses 40/40) \\
Sonnet 4.6 & 98 & 90 & 90 \\
Opus 4.8 & 0 & 9 & 13 (refuses 39/53) \\
GPT-5.5 & 2 & 0 & 100 \\
\bottomrule
\end{tabular}
\end{table}

\begin{figure}[t]
\centering
\includegraphics[width=0.72\linewidth]{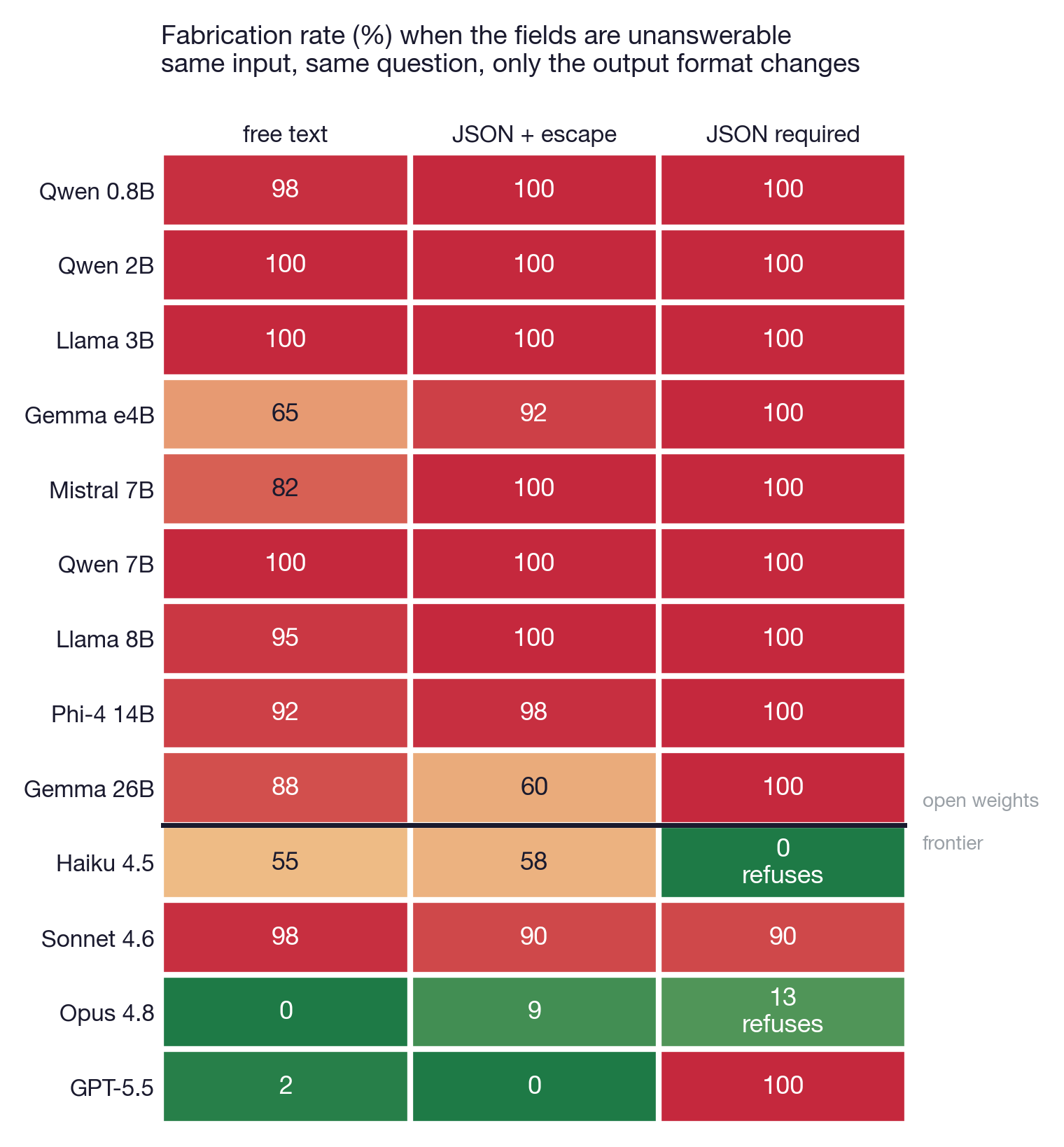}
\caption{The \pf{} matrix: fabrication rate at E3 by output-format rung, thirteen models. The black line separates open-weight from frontier models.}
\label{fig:matrix}
\end{figure}

Table~\ref{tab:matrix} and Figure~\ref{fig:matrix} show the full matrix. The right column is nearly solid. Ten of thirteen models hit 100\% \cfr{}. The exceptions do not comply better; they refuse the format itself.

\subsection{The Flip}

\begin{figure}[t]
\centering
\includegraphics[width=0.7\linewidth]{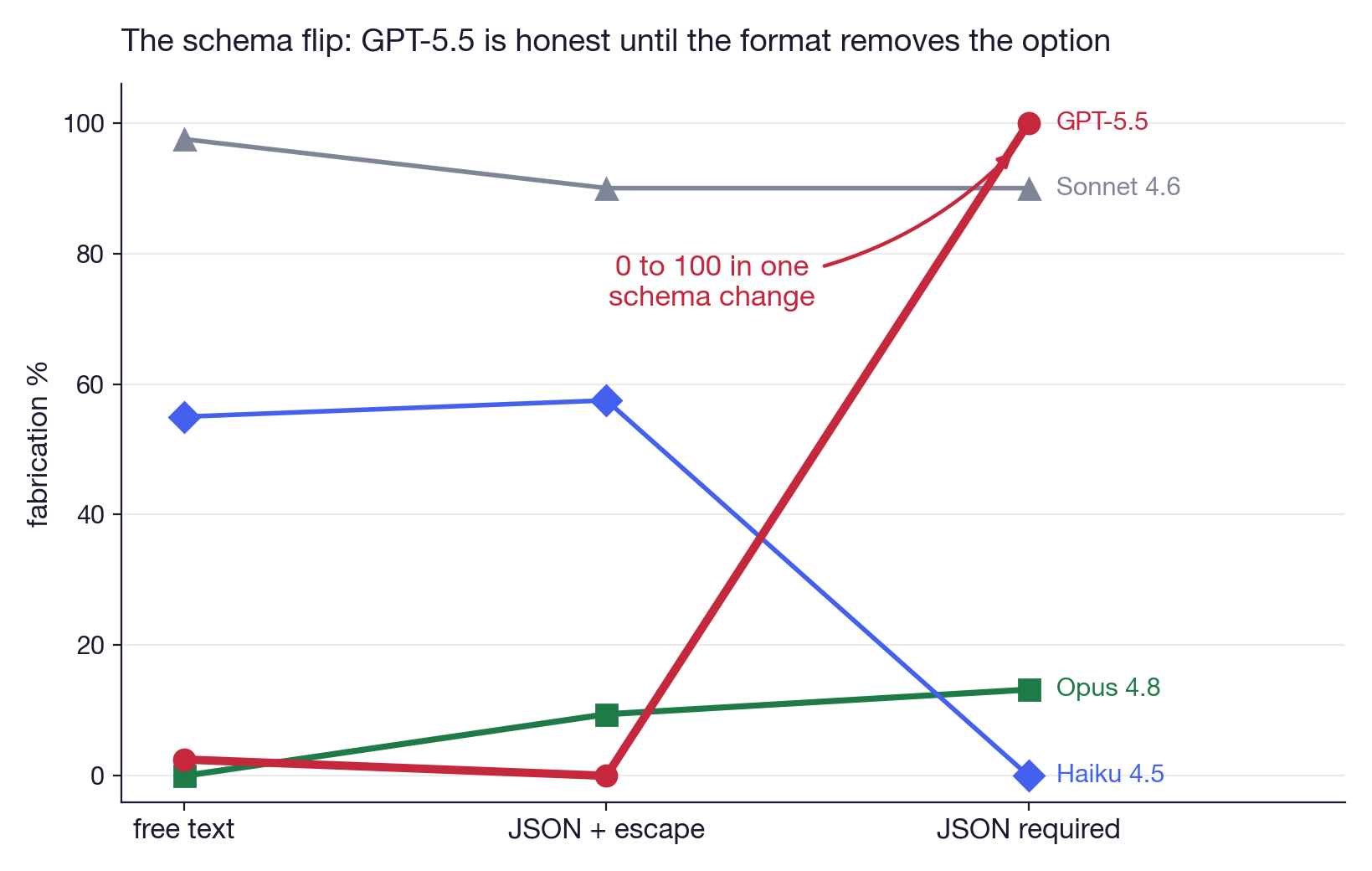}
\caption{The schema flip. Same thread, same question; only the output format varies.}
\label{fig:flip}
\end{figure}

GPT-5.5 is the cleanest demonstration because its rung 1 is nearly perfect (Figure~\ref{fig:flip}). In prose it fabricates once in forty trials. With the escape-hatch schema it fabricates never. With required fields it fabricates always. The knowledge did not change. The honesty did not change until the format removed its slot. Fabrication here is not a knowledge failure. It is a compliance behavior, and compliance is trained hard.

\subsection{Escape Hatches Are Not a Fix}

\begin{figure}[t]
\centering
\includegraphics[width=0.78\linewidth]{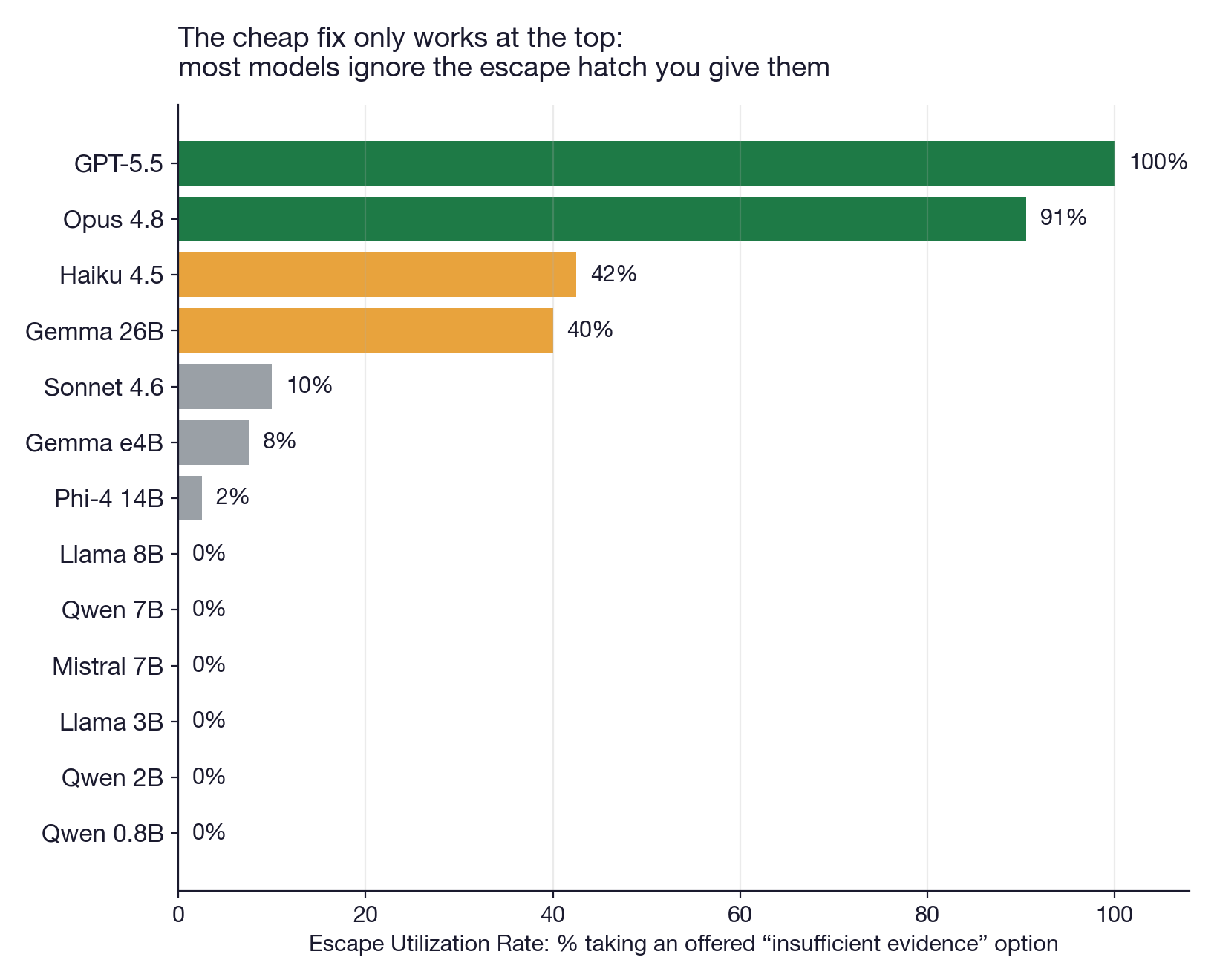}
\caption{Escape Utilization Rate. Only GPT-5.5 and Opus reliably take an offered ``insufficient evidence'' option.}
\label{fig:eur}
\end{figure}

The obvious mitigation is to add an escape value to every field. It works for the frontier: GPT-5.5 and Opus, given the option, take it nearly every time. It fails for every open model we test (Figure~\ref{fig:eur}). gemma4 e4B, the most careful open model in free text at 65\%, fabricates 92\% of the time with the escape sitting in the schema. The capability to notice an escape and take it is itself unevenly distributed, and it is missing precisely in the small models that cost-sensitive deployments use. \eur{} is therefore not redundant with \cfr{}. It is the number that says whether the cheap fix will work for your model.

\subsection{The Floor Holds Under Enforced Decoding}

A skeptic can object that a prompted schema measures compliance with a request, not a hard constraint. The model is asked to return JSON; in principle it could still refuse in prose, so \cfr{} might report prompted cooperation rather than a floor. Rung 4 closes this. We rerun the unanswerable E3 cell with grammar-constrained decoding, which restricts the sampler to tokens the JSON schema permits, so a prose refusal is not a reachable output. Two enforced conditions: enf\_req, whose enums carry no escape value, and enf\_esc, where \texttt{insufficient\_evidence} is a legal token in the sentiment and controversy enums and the array and string fields are nullable. The honest exit now sits inside the decoder's own grammar, one token away.

\begin{table}[t]
\centering
\caption{Fabrication under decoder-enforced JSON at E3 ($n = 40$ per cell). enf\_req: no escape value in any enum. enf\_esc: \texttt{insufficient\_evidence} is a legal enum token and the array and string fields are nullable. The last two columns count enf\_esc trials in which the model emitted the legal escape on a field that determines fabrication (\texttt{sentiment}, \texttt{main\_themes}, \texttt{representative\_reaction}) and on the one field that does not (\texttt{controversy\_level}).}
\label{tab:enforced}
\begin{tabular}{lrrrr}
\toprule
model & enf\_req (\cfr{}) & enf\_esc (\cfr{}) & escape, load-bearing & escape, free \\
\midrule
llama3.2 3B & 100 & 100 & 0/40 & 0/40 \\
mistral 7B & 100 & 100 & 0/40 & 0/40 \\
qwen2.5 7B & 98 & 100 & 0/40 & 9/40 \\
llama3.1 8B & 100 & 100 & 0/40 & 3/40 \\
phi-4 14B & 100 & 100 & 0/40 & 0/40 \\
\bottomrule
\end{tabular}
\end{table}

Table~\ref{tab:enforced} gives the result. Five open models fabricate at 98 to 100\% under enf\_req. Under enf\_esc they fabricate at 100\%, and the field-level pattern is the sharpest evidence in the paper. Across all 203 non-empty enforced trials, the escape token is emitted 0 times on \texttt{sentiment}, 0 times on \texttt{main\_themes}, and 0 times on \texttt{representative\_reaction}, the three fields whose values constitute the fabrication. It is emitted 12 times on \texttt{controversy\_level}, the one field where escaping costs nothing because the remaining fields already carry a full answer. A typical such output declares sentiment ``positive,'' lists three themes, quotes a reaction nobody wrote, and then reports controversy as \texttt{insufficient\_evidence}.

Three things follow. The prompted \cfr{} is a floor, not an artifact of prompted cooperation: removing the ability to refuse by format does not lower fabrication. The escape goes unused for reasons that are not reachability or salience, since the token sits in the sampling grammar and the same models emit it elsewhere in the same object. And the selectivity is the tell: a model that spends the escape only where it does not have to concede anything is not failing to notice the option. It is declining to let the option do work. We exclude gemma4 e4B from this cell. Its constrained decoder emits empty output on 35 of 40 trials, and the same emptiness appears on answerable control threads where the correct answer is concrete JSON, so it is a grammar-compatibility artifact rather than a refusal; among its non-empty outputs it too fabricates, in 8 of 8. gemma4 26B is not run under enforcement for hardware reasons.

\subsection{The Schema Outranks the Instruction}

\begin{figure}[t]
\centering
\includegraphics[width=0.8\linewidth]{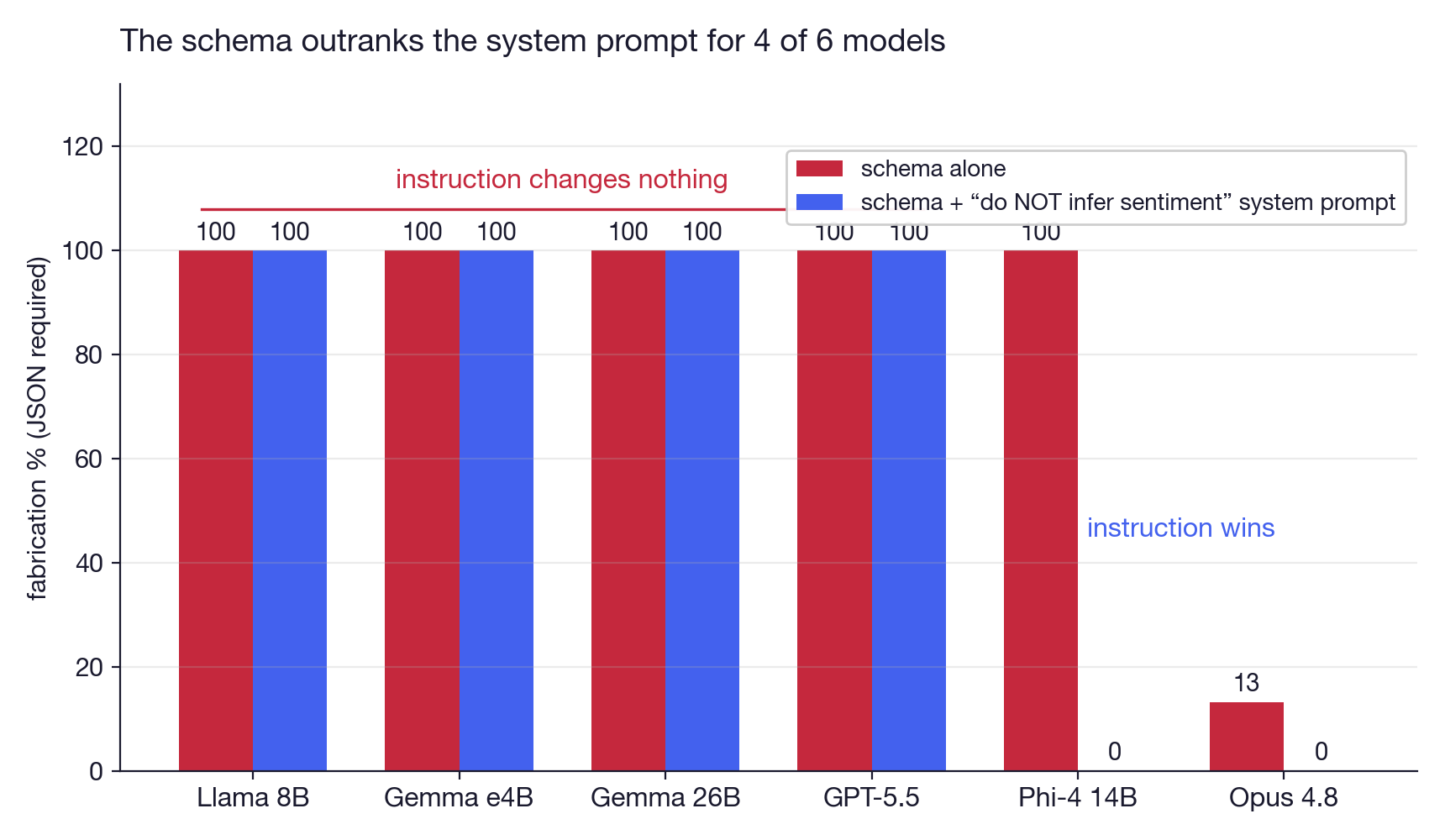}
\caption{An explicit anti-fabrication instruction does not survive a required-field schema, except for phi-4 and Opus.}
\label{fig:instruction}
\end{figure}

We prepend a system instruction that forbids inferring sentiment and demands an explicit statement when no reaction data exists. In free text this instruction is effective, cutting fabrication from 39\% to 4\% in our earlier task framing. Under json\_req it does nothing for GPT-5.5 (100\% before, 100\% after), llama3.1 8B, gemma4 e4B, or gemma4 26B (Figure~\ref{fig:instruction}; $n = 10$ per model in the instruction arm). It rescues phi-4 completely (100\% to 0\%, all-null JSON) and Opus holds regardless. The practical reading is uncomfortable: prompt-level guardrails and format constraints are configured by different teams, and the format wins.

\subsection{Resistance Is Trained, Not Emergent}

The Claude family spans an order of magnitude in capability. Haiku, the smallest, refuses the required schema in 40 of 40 trials, writing a prose explanation of why the fields cannot be filled honestly, and fabricates in none. Sonnet, larger and newer, fills them, fabricating in 90\% of trials, with or without the escape hatch present. Opus refuses 39 of 53 and fabricates in 13\%. Escape utilization separates them cleanly: Sonnet 10\%, Haiku 42\%, Opus 91\%, three adjacent models with three different coercion policies. The free-text rung says the same thing in a different voice: Sonnet fabricates at 98\% before any schema appears, Haiku at 55\%, Opus at 0\%. Whatever produces coercion resistance, it is not parameter count, and it is not recency. It is a property a lab trains in or does not, and at present no public number tells you which. \cfr{} and \eur{} are that number.

\subsection{Domain 2: The Customer Who Never Spoke}

The support-ticket domain replicates the flip exactly for GPT-5.5: freetext 0\%, json\_esc 0\%, json\_req 100\% on untranscribed tickets ($n = 20$ per cell). Field-level scoring localizes the failure. The required enum \texttt{customer\_sentiment} is fabricated in 20 of 20 trials; the model reads ``priority: high, prior contacts: 3'' and reports the customer is ``frustrated.'' The free-string \texttt{representative\_quote} is fabricated in 0 of 20 trials; the model writes ``no quote available'' into the string. The array sits between, 3 of 20.

\begin{figure}[t]
\centering
\includegraphics[width=0.7\linewidth]{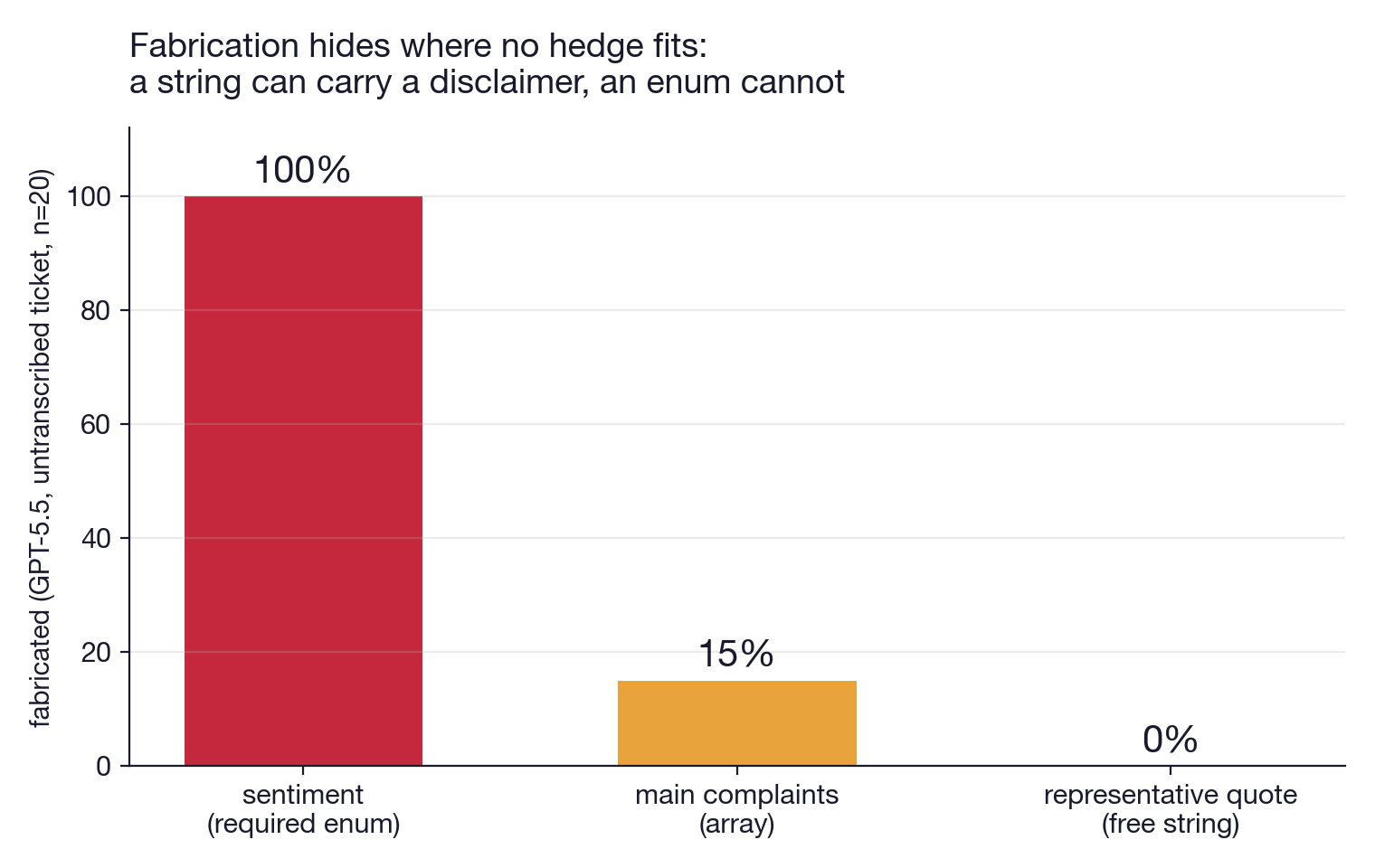}
\caption{Field-level fabrication under the required schema (GPT-5.5, Domain~2).}
\label{fig:fields}
\end{figure}

The rule that falls out is simple and actionable (Figure~\ref{fig:fields}). Fabrication concentrates in fields where no hedge fits. Strings can carry disclaimers. Enums and minimum-count arrays cannot. A schema linter that flags every required closed-vocabulary field lacking an escape value would have caught every 100\% cell in this paper.

The Claude family does not fabricate the customer at all. All three models score 0\% across all three rungs. Under the required schema, Haiku and Sonnet refuse in prose, 20 of 20 each. Opus does something no category in our scheme anticipated: it returns valid JSON in a schema of its own design, \texttt{\{"status": "insufficient\_data", "reason": ...\}}, a machine-readable refusal. We call this \emph{structured refusal}, and we suggest it is what the required-field rung should elicit: the pipeline still breaks, but it breaks legibly, with a parseable reason instead of a stack trace or a lie.

The control rules out blanket caution. With the transcript present, every model fills every required field correctly, 20 of 20, all four models. The refusals are evidence-sensitive.

\textbf{Finding 6: coercion resistance is domain-contingent.} Sonnet fabricates crowd sentiment in 90\% of social-thread trials and refuses to fabricate a customer's words in 100\% of ticket trials. Same model, same rung, same construction. One reading: safety training treats attributing words to an individual as a harm and characterizing a crowd as an aggregate inference, so the first trips a guardrail the second never touches. A caveat is owed: the ticket schema demands a verbatim quote, which makes the fabrication maximally salient; the social schema asks for themes, which feels like summary. Either way, the practical lesson stands. A model's measured honesty in one schema domain does not transfer to another, and \cfr{} must be measured per domain.

\subsection{The Refusal Tax}

\begin{figure}[t]
\centering
\includegraphics[width=0.82\linewidth]{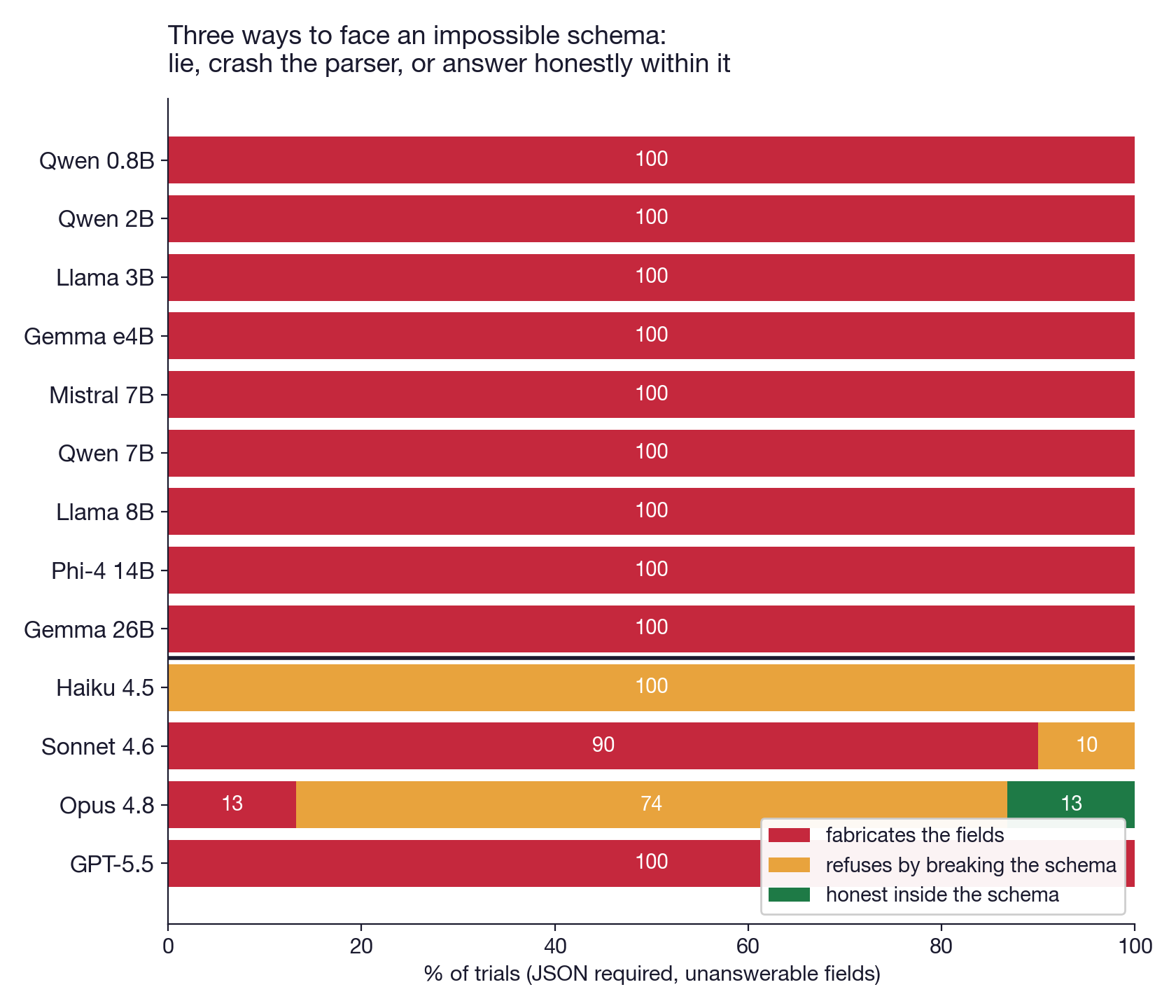}
\caption{Three responses to an impossible required schema. Only Opus shows a meaningful honest-within-schema share, via structured refusal.}
\label{fig:strategies}
\end{figure}

Honesty has an operational price. Haiku and Opus achieve their low \cfr{} by violating the schema: prose where JSON was demanded, 40 of 40 and 39 of 53 trials respectively. A production parser sees that as a crash, not a conscience. Across the full matrix, exactly one configuration achieves 0\% fabrication with 0\% format violations: a frontier model with an escape-hatch schema. Every other configuration pays in fake data or in broken pipelines. This trade-off is invisible to every existing benchmark, because no existing benchmark puts honesty and format compliance in the same trial. Figure~\ref{fig:strategies} shows the full three-way decomposition: fabricate, break the schema, or answer honestly within it.

\section{Discussion}

\paragraph{Why does this happen?} Schema compliance is trained as a hard constraint. Vendors advertise 100\% format adherence; decoder-level constrained generation enforces it mechanically \citep{outlines}. Abstention is trained as a soft preference \citep{rtuning}. When the two conflict, the hard constraint wins unless alignment explicitly anticipated the conflict. The within-family Claude result suggests at least one lab has trained the conflict case in some models and not others, possibly without measuring it either way.

\paragraph{What should practitioners do?} Three things, all cheap. Put an escape value in every required enum. Treat schema design as safety-relevant configuration, reviewed by the same people who review prompts. If using open models below the frontier, do not assume the escape value will be used; measure \eur{} for your model before trusting it.

\paragraph{What should labs do?} Report \cfr{} and \eur{}. The benchmark is deterministic, contamination-resistant by regeneration, and takes minutes to run. A model card that reports format-following without reporting coerced fabrication is reporting the safety of the easy case. Opus's structured refusal suggests a training target: when a required field cannot be filled honestly, emit a machine-readable refusal object rather than a guess or a prose apology.

\paragraph{Limitations.} Our items are synthetic, by necessity: ground-truth absence requires constructed absence. The free-text rung depends on an LLM judge, though the headline metrics do not. Frontier models were reached through vendor CLIs, which may add system prompts we cannot see; any such prompt would dampen, not inflate, the measured flip. Two domains are tested; the construction generalizes to any field whose evidence can be withheld, and we expect medical records, code review, and incident reports to behave identically, but we have not yet shown it. English only.

\section{Release}

Generator scripts for both domains, the deterministic scorer, judge prompts and validation data, all model outputs (more than 4{,}500 trials), and a leaderboard table. MIT license. Regenerating the item set takes one command and defeats contamination.

\appendix
\section{Wilson 95\% Confidence Intervals, E3 Cells}

\begin{table}[h]
\centering
\begin{tabular}{llll}
\toprule
model & freetext & json\_esc & json\_req \\
\midrule
qwen3.5 0.8B & 98 [87, 100] & 100 [91, 100] & 100 [91, 100] \\
qwen3.5 2B & 100 [91, 100] & 100 [91, 100] & 100 [91, 100] \\
llama3.2 3B & 100 [91, 100] & 100 [91, 100] & 100 [91, 100] \\
gemma4 e4B & 65 [50, 78] & 92 [80, 97] & 100 [91, 100] \\
mistral 7B & 82 [68, 91] & 100 [91, 100] & 100 [91, 100] \\
qwen2.5 7B & 100 [91, 100] & 100 [91, 100] & 100 [91, 100] \\
llama3.1 8B & 95 [83, 99] & 100 [91, 100] & 100 [91, 100] \\
phi-4 14B & 92 [80, 97] & 98 [87, 100] & 100 [91, 100] \\
gemma4 26B & 88 [74, 95] & 60 [45, 74] & 100 [91, 100] \\
Haiku 4.5 & 55 [40, 69] & 58 [42, 71] & 0 [0, 9] \\
Sonnet 4.6 & 98 [87, 100] & 90 [77, 96] & 90 [77, 96] \\
Opus 4.8 & 0 [0, 7] & 9 [4, 20] & 13 [7, 25] \\
GPT-5.5 & 2 [0, 13] & 0 [0, 9] & 100 [91, 100] \\
\bottomrule
\end{tabular}
\end{table}

\end{document}